\documentclass[conference,twocolumn]{IEEEtran}

\usepackage[utf8]{inputenc}
\usepackage[T1]{fontenc}
\usepackage{amsmath,amssymb,amsfonts}
\usepackage{graphicx}
\usepackage{booktabs}
\usepackage{multirow}
\usepackage{hyperref}
\usepackage{cleveref}
\usepackage{xcolor}
\usepackage{balance}
\usepackage{microtype}
\usepackage{tabularx}
\usepackage{array}
\usepackage{makecell}
\usepackage{float}
\usepackage{orcidlink}
\usepackage{placeins}

\newcommand{\kaggleurl}{\href{https://kaggle.com/datasets/mfazrinizar/fetal-head-segmentation-yolo-splitted}{Dataset}}
\newcommand{\githuburl}{\href{https://github.com/mfazrinizar/fetal-head-segmentation}{Code}}

\hypersetup{
  colorlinks=true,
  linkcolor=blue,
  citecolor=blue,
  urlcolor=blue,
  hypertexnames=false
}

\usepackage{enumitem}
\setlist{nosep}

\newcommand{\doi}[1]{\href{https://doi.org/#1}{DOI:~#1}}

\title{Domain-Guided YOLO26 with Composite BCE-Dice-Lov\'{a}sz Loss\\for Multi-Class Fetal Head Ultrasound Segmentation}

\author{%
  \IEEEauthorblockN{M.~Fazri Nizar~\orcidlink{0009-0002-8330-8520}}\\
  \IEEEauthorblockA{%
    Informatics Engineering, Faculty of Computer Science\\
    Sriwijaya University\\
    \texttt{mfazrinizar@gmail.com},
    \texttt{09021282328053@student.unsri.ac.id}%
  }%
}

\begin{document}
\maketitle

\begin{abstract}
Segmenting fetal head structures from prenatal ultrasound remains a practical bottleneck in obstetric imaging.
The current state-of-the-art baseline, proposed alongside the published dataset,
adapts the Segment Anything Model with per-class Dice and Lov\'{a}sz losses but still depends on
bounding-box prompts at test time.
We build a prompt-free pipeline on top of YOLO26-Seg that jointly detects and segments three
structures, Brain, Cavum Septi Pellucidi (CSP), and Lateral Ventricles (LV), in a single forward pass.
Three modifications are central to our approach:
(i)~a composite BCE-Dice-Lov\'{a}sz segmentation loss with inverse-frequency class weighting, injected into
the YOLO26 training loop via runtime monkey-patching;
(ii)~domain-guided copy-paste augmentation that transplants minority-class structures while respecting their
anatomical location relative to the brain boundary; and
(iii)~inter-patient stratified splitting to prevent data leakage.
On 575 held-out test images, the composite loss variant reaches a mean Dice coefficient of 0.9253,
exceeding the baseline (0.9012) by 2.68 percentage points, despite reporting over three foreground
classes only, whereas the baseline's reported mean includes the easy background class.
We further ablate each component and discuss annotation-quality and class-imbalance effects on CSP and LV
performance.
\end{abstract}


\section{Introduction}
\label{sec:introduction}

Routine prenatal ultrasound relies on measurements such as head circumference (HC), biparietal diameter (BPD),
and ventricular width to track fetal development and flag anomalies~\cite{hadlock1985}.
Extracting these measurements requires delineating the fetal brain boundary, the Cavum Septi Pellucidi (CSP),
and the Lateral Ventricles (LV), three structures whose clinical relevance is well
established~\cite{cardoza1988,jou1998}. In practice, manual annotation is slow, subjective, and exhibits
considerable inter-observer variability~\cite{sarris2012}.

Over the past decade, convolutional and attention-based architectures, U-Net~\cite{ronneberger2015},
UNet++~\cite{zhou2018}, SegFormer~\cite{xie2021}, DeepLabV3+~\cite{chen2018deeplabv3p}, have steadily
narrowed the gap between automated and expert segmentation, as reviewed comprehensively by Fiorentino
et~al.~\cite{fiorentino2023}. More recently, Alzubaidi et~al.~\cite{alzubaidi2024} fine-tuned the Segment
Anything Model (SAM)~\cite{kirillov2023} with a weighted Dice-Lov\'{a}sz objective, producing FetSAM.
While FetSAM reports a mean DSC of 0.9012 across four classes (including background), it requires a
bounding-box prompt for every input, which in turn requires either a separate detector or manual interaction.

The YOLO family~\cite{redmon2016,jocher2023}, whose evolution from YOLOv1 through YOLOv8 is surveyed by
Terven et~al.~\cite{terven2023}, offers single-pass detection and instance segmentation without auxiliary
prompts. YOLO26~\cite{sapkota2025}, the latest release, introduces C2PSA attention blocks, building on the
self-attention mechanism of Vaswani et~al.~\cite{vaswani2017}, and a hybrid Muon-SGD optimizer (MuSGD)
that orthogonalizes gradient updates via Newton--Schulz iterations. Meanwhile, Ghiasi
et~al.~\cite{ghiasi2021} showed that copy-paste augmentation can meaningfully boost minority-class recall
in instance segmentation, and Athalye and Arnaout~\cite{athalye2023} demonstrated domain-guided copy-paste
specifically for medical ultrasound imaging.

This paper makes four contributions:
\begin{enumerate}
  \item \textbf{Composite BCE-Dice-Lov\'{a}sz loss.} A three-term segmentation loss with inverse-frequency
    class weights, integrated into YOLO26 via monkey-patching rather than source modification.
  \item \textbf{Domain-guided augmentation.} An augmentation scheme that copies CSP/LV regions from donor
    images and pastes them into brain-only acceptor images at anatomically consistent locations.
  \item \textbf{Inter-patient splitting.} Patient-level partitioning of the HC18-derived dataset, published
    publicly, that prevents data leakage.
  \item \textbf{Ablation and comparison.} Evaluation across three configurations with per-class analysis,
    compared against FetSAM and twelve other architectures from~\cite{alzubaidi2024}.
\end{enumerate}

\section{Materials and Methods}
\label{sec:methods}

\subsection{Dataset}
\label{sec:dataset}

The images originate from the HC18 Grand Challenge~\cite{vandenheuvel2018} and the large-scale fetal
head biometry annotation dataset of Alzubaidi et~al.~\cite{alzubaidi2023data}, which provides 3{,}832
high-resolution ultrasound images (959$\times$661\,px) with polygon masks verified by a senior attending
physician and a radiologic technologist (ICC $\geq$ 0.859). Each 2D B-mode frame is annotated with
polygon masks for up to three structures:
\begin{itemize}
  \item \textbf{Brain (Class~0):} The full cranial boundary, the largest and easiest to delineate.
  \item \textbf{CSP (Class~1):} A small midline cavity whose apparent size varies with gestational
    age~\cite{jou1998}.
  \item \textbf{LV (Class~2):} Paired fluid-filled chambers; enlargement beyond 10\,mm suggests
    ventriculomegaly~\cite{cardoza1988}.
\end{itemize}

\begin{table}[!htb]
\centering
\caption{Instance counts after inter-patient splitting.}
\label{tab:dataset}
\begin{tabular}{lrrrrr}
\toprule
Split   & Images & Brain & CSP   & LV    & Total \\
\midrule
Train   & 2{,}654 & 2{,}697 & 921   & 1{,}052 & 4{,}670 \\
Val     & 603   & 611   & 190   & 212   & 1{,}013 \\
Test    & 575   & 568   & 172   & 212   & 952   \\
\midrule
\textbf{All} & \textbf{3{,}832} & \textbf{3{,}792} & \textbf{1{,}283} & \textbf{1{,}476} & \textbf{6{,}551} \\
\bottomrule
\end{tabular}
\end{table}

\begin{figure}[!htb]
\centering
\includegraphics[width=\columnwidth]{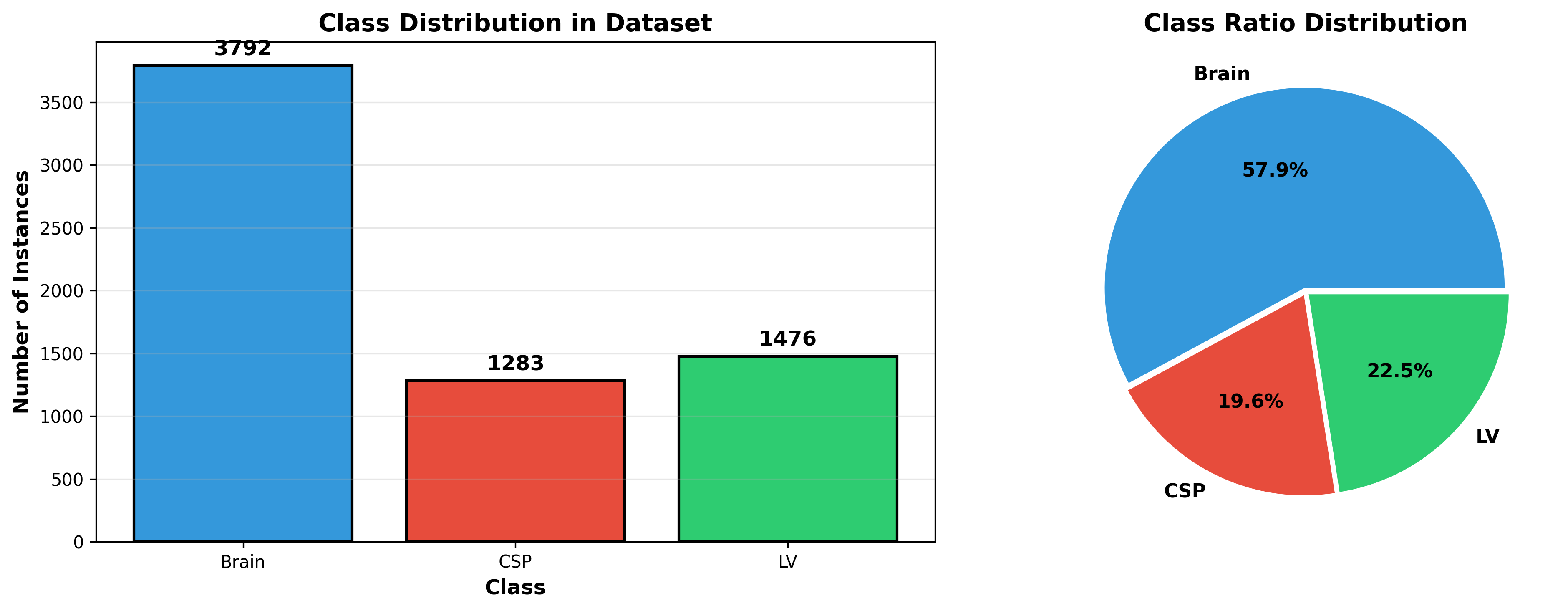}
\caption{Class distribution and imbalance ratio across the full dataset.}
\label{fig:class_dist}
\end{figure}

\begin{figure}[!htb]
\centering
\includegraphics[width=\columnwidth]{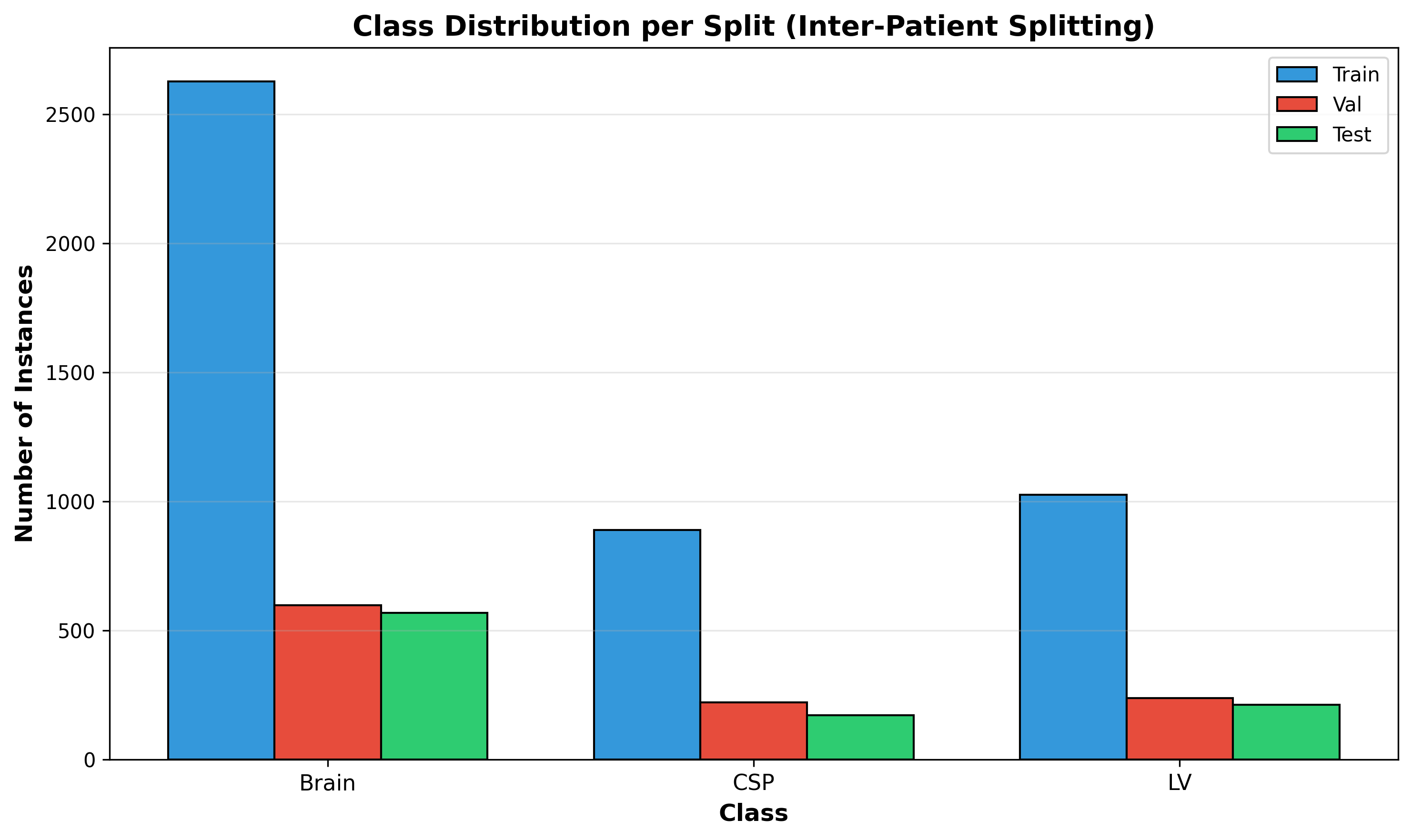}
\caption{Per-split class counts, showing that the stratified split preserves proportional representation.}
\label{fig:class_splits}
\end{figure}

\Cref{tab:dataset} summarizes the per-split instance counts. Brain accounts for 57.9\% of all instances;
the Brain-to-CSP ratio is roughly 3:1. Not every image contains all three structures, many show only the
brain contour. \Cref{fig:class_dist} visualizes the class distribution and imbalance ratio, while
\Cref{fig:class_splits} confirms that the stratified split preserves proportional representation.

\subsection{Inter-Patient Splitting}
\label{sec:splitting}

In the HC18 dataset, a single patient can contribute multiple frames. A na\"{i}ve random split therefore
risks placing frames from the same patient in both training and test partitions, inflating reported
scores~\cite{rieke2020}. This type of data leakage is a well-documented pitfall in medical imaging
studies~\cite{fiorentino2023}. We group all frames by patient ID (extracted from the filename convention),
then allocate patients, not images, to train/val/test at a 70/15/15 ratio with approximate class
stratification. The resulting splits are publicly released\footnote{\kaggleurl} in YOLO polygon format.

\subsection{YOLO26-Seg Architecture}
\label{sec:architecture}

YOLO26-Seg~\cite{sapkota2025} performs detection and instance segmentation in one pass. Its backbone stacks
Conv layers (stride~2) with batch normalization~\cite{ioffe2015}, C3k2 CSP bottleneck blocks~\cite{wang2020},
an SPPF module with three cascaded $5\times5$ max-pools, and a C2PSA block. C2PSA is the distinguishing
element: it splits channels via a $1\times1$ convolution, feeds one branch through PSABlock modules that
apply multi-head self-attention~\cite{vaswani2017} (heads $= c/64$) and a feed-forward network, then
concatenates and projects back. This gives the network global context for relating spatially distant
structures, useful when the model must learn that CSP and LV always appear inside the brain. Deep residual
connections~\cite{he2016resnet} within the C3k2 blocks facilitate gradient flow through the network.

The neck fuses features bidirectionally via Feature Pyramid Network (FPN)~\cite{lin2017fpn} top-down and
Path Aggregation Network (PAN)~\cite{liu2018panet} bottom-up pathways through C3k2 blocks at P3--P5.
The Segment26 head predicts boxes, class logits, and 32-dim mask prototype coefficients; the final mask
is a linear combination of learned prototypes, following the prototype-based paradigm introduced by
YOLACT~\cite{bolya2019} and later adopted by Mask R-CNN variants~\cite{he2017maskrcnn}.
\Cref{tab:variants} lists the two variants used in this study along with their inference speed, parameter
count, and FLOPs.

\begin{table}[!htb]
\centering
\caption{Model variants used in this work.}
\label{tab:variants}
\begin{tabular}{lcccc}
\toprule
Variant & \makecell{Speed CPU\\ONNX (ms)} & \makecell{Speed GPU\\T4 TRT10 (ms)} & Params & FLOPs \\
\midrule
YOLO26N-Seg & 53.3\,$\pm$\,0.5 & 2.1\,$\pm$\,0.0 & 2.7\,M & 9.1 \\
YOLO26L-Seg & 387.0\,$\pm$\,3.7 & 8.0\,$\pm$\,0.1 & 28.0\,M & 139.8 \\
\bottomrule
\end{tabular}
\end{table}

\subsection{Composite BCE-Dice-Lov\'{a}sz Loss}
\label{sec:loss}

YOLO26 ships with a \texttt{BCEDiceLoss} for the segmentation head. We replace it at runtime with a
three-term composite that adds a Lov\'{a}sz Hinge component and per-class weighting.

\smallskip\noindent\textbf{Binary Cross-Entropy.} Given $N$ pixels with ground truth $y_i\in\{0,1\}$
and predicted probability $\hat{y}_i$:
\begin{equation}
\mathcal{L}_{\text{BCE}} = -\frac{1}{N}\sum_{i=1}^{N}\bigl[y_i\ln\hat{y}_i + (1-y_i)\ln(1-\hat{y}_i)\bigr]
\label{eq:bce}
\end{equation}

\smallskip\noindent\textbf{Dice Loss.} With smoothing constant $\epsilon=1$:
\begin{equation}
\mathcal{L}_{\text{Dice}} = 1 - \frac{2\sum_i\hat{y}_i y_i + \epsilon}{\sum_i\hat{y}_i + \sum_i y_i + \epsilon}
\label{eq:dice}
\end{equation}

\smallskip\noindent\textbf{Lov\'{a}sz Hinge Loss}~\cite{berman2018}. For a binary mask of class~$c$,
let the per-pixel hinge errors $m_i = 1-(2y_i-1)\hat{y}_i$ be sorted in decreasing order. The loss is
the inner product of the positive errors with the discrete gradient $\Delta J$ of the Jaccard index
evaluated along that ordering:
\begin{equation}
\mathcal{L}_{\text{Lov\'{a}sz}} = \sum_i \max(0,\,m_i)\,\Delta J_i
\label{eq:lovasz}
\end{equation}
where $\Delta J_i = J(\{1,\dots,i\}) - J(\{1,\dots,i-1\})$, and $J$ denotes the Jaccard set function
over the sorted error positions. Whereas Dice loss optimises the F1 surrogate, Lov\'{a}sz provides a tight
convex extension of the IoU metric itself~\cite{berman2018}. This is related to, but distinct from, focal
loss~\cite{lin2017focal}, which addresses class imbalance by down-weighting easy examples rather than
directly optimising the set-based overlap metric. We assign inverse-frequency class weights
(\Cref{tab:weights}) to the Dice and Lov\'{a}sz terms so that minority classes receive stronger gradients.

\begin{table}[!htb]
\centering
\caption{Inverse-frequency class weights used in Dice and Lov\'{a}sz terms.}
\label{tab:weights}
\begin{tabular}{lcrr}
\toprule
Class & Instances & Weight & Normalised \\
\midrule
Brain & 3{,}792 (57.9\%) & 0.1 & 0.059 \\
CSP   & 1{,}283 (19.6\%) & 0.9 & 0.529 \\
LV    & 1{,}476 (22.5\%) & 0.7 & 0.412 \\
\bottomrule
\end{tabular}
\end{table}

The composite loss sums the three terms:
\begin{equation}
\mathcal{L} = 0.25\,\mathcal{L}_{\text{BCE}} + 0.50\,\mathcal{L}_{\text{Dice}} + 0.25\,\mathcal{L}_{\text{Lov\'{a}sz}}
\label{eq:composite}
\end{equation}

Dice receives the largest coefficient because DSC is the primary evaluation metric; BCE supplies stable
per-pixel gradients during early training; and the Lov\'{a}sz term sharpens boundary predictions through
direct IoU optimisation.

\smallskip\noindent\textbf{Integration.} Rather than forking the YOLO26 codebase, we monkey-patch
\texttt{ultralytics.utils.loss.BCEDiceLoss} with our \texttt{FetSAMBCEDiceLovaszLoss} class before model
instantiation. This leaves all other training machinery, data loading, augmentation, scheduling,
checkpointing, untouched. Shape mismatches between multi-scale mask predictions and targets are resolved
via bilinear interpolation inside the patched loss.

\subsection{Automatic Optimizer Selection (MuSGD)}
\label{sec:optimizer}

YOLO26 introduces an automatic optimizer selection mechanism. When the optimizer argument is set to
\texttt{auto}, the framework estimates the total number of training iterations
$I = E\times\lceil N_{\text{train}}/B\rceil$ (epochs $\times$ steps per epoch) and branches:
\begin{equation}
\text{optimizer} = \begin{cases}
\text{MuSGD},\;\text{lr}=0.01,\;\mu=0.9 & \text{if } I > 10{,}000 \\
\text{AdamW},\;\text{lr}=\tfrac{0.01}{4+n_c},\;\mu=0.9 & \text{otherwise}
\end{cases}
\label{eq:optimizer}
\end{equation}
where $n_c$ is the number of classes. AdamW~\cite{loshchilov2019adamw} decouples weight decay from the
gradient update, which improves generalisation over the classical Adam formulation. MuSGD is a hybrid
optimizer that blends Muon updates (momentum with Newton--Schulz orthogonalisation of the gradient matrix)
and classical SGD with Nesterov momentum. For 2-D+ parameter tensors, the Muon branch computes an
approximate orthogonal factorisation $G\mapsto UV^{\top}$ via five Newton--Schulz iterations, which acts
as a form of gradient preconditioning. The mixing coefficients are fixed at
$w_{\text{muon}}=0.2$, $w_{\text{sgd}}=1.0$.

In our experiments, the Domain-Guided run ($300\times\lceil 2654/16\rceil\approx 49{,}762$ iterations)
triggered MuSGD, while the FetSAM Loss run ($300\times\lceil 2654/128\rceil\approx 6{,}220$ iterations)
fell below the threshold and used AdamW with a fitted learning rate of 0.001429. The Baseline used an
explicitly specified AdamW.

\subsection{Domain-Guided Copy-Paste Augmentation}
\label{sec:augmentation}

Standard geometric transforms (flips, rotations, scaling) do not synthesise new minority-class instances.
Following the copy-paste framework of Ghiasi et~al.~\cite{ghiasi2021} and the domain-guided principles
of Athalye and Arnaout~\cite{athalye2023}, we implement a domain-guided variant of copy-paste augmentation
that respects anatomical layout:

\begin{enumerate}
  \item \textbf{Categorise} training images into \emph{acceptors} (Brain-only), \emph{CSP donors}, and
    \emph{LV donors}.
  \item \textbf{Measure} the donor structure's centroid offset from the brain centroid:
    $(\Delta x,\,\Delta y)$.
  \item \textbf{Paste} the structure into the acceptor at the same relative offset inside the acceptor's
    brain region, requiring $\geq70\%$ spatial overlap with the brain mask.
  \item \textbf{Blend} with $\alpha=0.95$ to preserve texture continuity.
  \item \textbf{Re-extract} polygon labels via contour detection on the composited mask.
\end{enumerate}

The procedure runs once as an offline preprocessing step (flagged by a marker file to prevent
re-application). It specifically increases CSP and LV representation without generating anatomically
impossible composites, unlike Mosaic or MixUp, which we therefore disable.

\subsection{Training Configuration}
\label{sec:training}

\Cref{tab:hyperparams} lists all hyperparameters for the three experiments, and \Cref{tab:training_time}
reports the corresponding wall-clock training times.

\begin{table}[!htb]
\centering
\caption{Hyperparameters per experiment.}
\label{tab:hyperparams}
\setlength{\tabcolsep}{2.5pt}
\begin{tabular}{lccc}
\toprule
Parameter & Baseline & Domain-Guided & FetSAM Loss \\
\midrule
Architecture & YOLO26N-Seg & YOLO26L-Seg & YOLO26N-Seg \\
Params & 2.7\,M & 28.0\,M & 2.7\,M \\
Input size & 640 & 800 & 640 \\
Batch size & 64 & 16 & 128 \\
Epochs & 180 (early stop) & 300 (3 sess.) & 300 \\
Optimizer & AdamW & Auto$\to$MuSGD & Auto$\to$AdamW \\
Effective lr & 0.0001 & 0.01 & 0.001429 \\
Weight decay & 0.001 & 0.0005 & 0.001 \\
Patience & 20 & 50 & 100 \\
Seg.\ loss & BCEDice & BCE-Dice-Lov. & BCE-Dice-Lov. \\
Augmentation & Standard & Domain-guided & Standard \\
Pretrained & Yes & Yes & No \\
Mosaic/MixUp & Off/Off & Off/Off & Off/Off \\
AMP & On & On & On \\
GPUs & 2$\times$T4 & 2$\times$T4 & 2$\times$T4 \\
Seed & 42 & 42 & 42 \\
\bottomrule
\end{tabular}
\end{table}

\begin{table}[!htb]
\centering
\caption{Wall-clock training times on T4\,$\times$2.}
\label{tab:training_time}
\begin{tabular}{lccrr}
\toprule
Experiment & Sessions & Epochs & Total time & Per epoch \\
\midrule
Baseline & 1 & 180 & 2.45\,h & 48.8\,s \\
Domain-Guided & 3 & 300 & 30.88\,h & 368.3\,s \\
FetSAM Loss & 1 & 300 & 4.04\,h & 48.3\,s \\
\bottomrule
\end{tabular}
\end{table}

As shown in \Cref{tab:training_time}, the Domain-Guided run spans three GPU training sessions
(112 + 116 + 72 epochs) because of the 12-hour per-session limit. Its longer per-epoch time (368\,s vs.\ 48\,s) reflects the 10$\times$ parameter count
and the higher input resolution. All runs use Automatic Mixed Precision (AMP)~\cite{micikevicius2018},
which stores activations in FP16 while accumulating gradients in FP32, reducing memory usage and
accelerating training on T4 GPUs. We disable Mosaic and MixUp throughout: stitching ultrasound frames from
different patients into a single tile produces anatomically incoherent images that harm rather than help
medical segmentation.

\subsection{Evaluation Protocol}
\label{sec:evaluation}

\noindent\textbf{Dice Similarity Coefficient (DSC)}~\cite{dice1945}:
\begin{equation}
\text{DSC}(A,B) = \frac{2|A\cap B|}{|A|+|B|}
\label{eq:dsc}
\end{equation}

\noindent\textbf{Intersection over Union (IoU)}:
\begin{equation}
\text{IoU}(A,B) = \frac{|A\cap B|}{|A\cup B|}
\label{eq:iou}
\end{equation}

\noindent\textbf{Mean Average Precision (mAP)} follows the COCO protocol~\cite{lin2014coco}. At each of
ten IoU thresholds (0.50, 0.55, \ldots, 0.95), predictions are matched to ground truths independently, a
precision--recall curve is built, its envelope is sampled at 101 recall points, and AP is the area under
the interpolated curve:
\begin{equation}
\text{AP} = \int_0^1 p_{\text{env}}(r)\,dr \approx \frac{1}{101}\sum_{k=0}^{100}p_{\text{env}}\!\left(\frac{k}{100}\right)
\label{eq:ap}
\end{equation}

mAP@50 uses only the 0.50 threshold; mAP@50-95 averages over all ten. Per-class Precision, Recall,
Specificity, and F1 are computed at IoU $\geq$ 0.5, then macro-averaged for overall scores.

\section{Results}
\label{sec:results}

\subsection{Overall Performance}

\begin{table}[!htb]
\centering
\caption{Test-set results (575 images). Best values in bold.}
\label{tab:overall}
\begin{tabular}{lccc}
\toprule
Metric & Baseline & Domain-Guided & FetSAM Loss \\
\midrule
mAP@50    & 0.7423 & \textbf{0.8657} & 0.7700 \\
mAP@50-95 & 0.4736 & \textbf{0.5462} & 0.5046 \\
mDSC      & 0.9229 & 0.9208          & \textbf{0.9253} \\
mIoU      & 0.8661 & 0.8624          & \textbf{0.8697} \\
Precision & 0.7393 & \textbf{0.8244} & 0.7592 \\
Recall    & 0.7810 & \textbf{0.8775} & 0.8178 \\
F1        & 0.7573 & \textbf{0.8467} & 0.7839 \\
Specificity & 0.8952 & \textbf{0.9444} & 0.9242 \\
Accuracy  & 0.7196 & \textbf{0.8204} & 0.7572 \\
\bottomrule
\end{tabular}
\end{table}

\begin{figure}[!htb]
\centering
\includegraphics[width=\columnwidth]{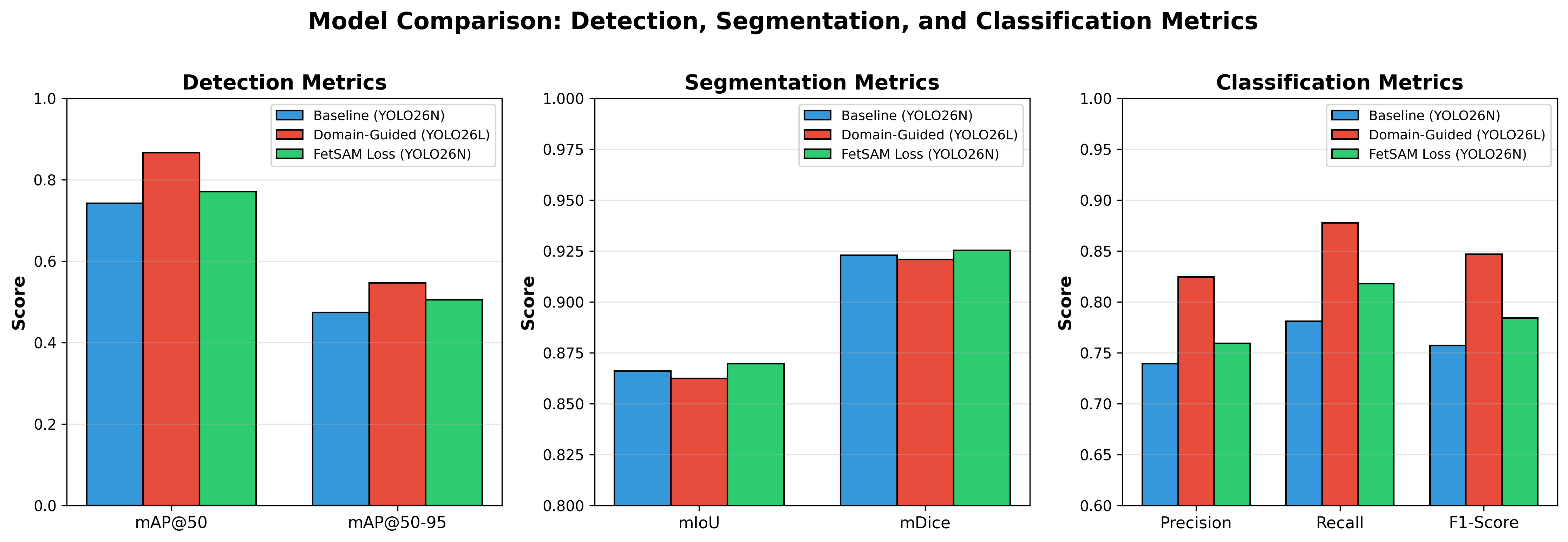}
\caption{Grouped bar chart of detection, segmentation, and classification metrics.}
\label{fig:model_comparison}
\end{figure}

\Cref{tab:overall} presents the aggregate test-set metrics. The Domain-Guided configuration leads on
every detection and classification metric (mAP@50 $+$16.6\,pp over baseline). In segmentation overlap,
however, the lighter FetSAM Loss variant edges ahead (mDSC 0.9253 vs.\ 0.9208), indicating that the
composite loss alone accounts for most of the segmentation gain, while the larger model and augmentation
chiefly improve detection. \Cref{fig:model_comparison} visualizes these differences as a grouped bar chart.

\subsection{Per-Class Breakdown}

\begin{table}[!htb]
\centering
\caption{Per-class results, Baseline (YOLO26N, standard BCEDice).}
\label{tab:perclass_baseline}
\setlength{\tabcolsep}{3pt}
\begin{tabular}{lcccccrrr}
\toprule
Class & Prec. & Rec. & F1 & IoU & DSC & TP & FP & FN \\
\midrule
Brain & 0.939 & 0.982 & 0.960 & 0.943 & 0.971 & 558 & 36 & 10 \\
CSP   & 0.624 & 0.762 & 0.686 & 0.704 & 0.823 & 131 & 79 & 41 \\
LV    & 0.655 & 0.599 & 0.626 & 0.695 & 0.816 & 127 & 67 & 85 \\
\bottomrule
\end{tabular}
\end{table}

\begin{table}[!htb]
\centering
\caption{Per-class results, Domain-Guided (YOLO26L, BCE-Dice-Lov\'{a}sz + domain aug.).}
\label{tab:perclass_dg}
\setlength{\tabcolsep}{3pt}
\begin{tabular}{lcccccrrr}
\toprule
Class & Prec. & Rec. & F1 & IoU & DSC & TP & FP & FN \\
\midrule
Brain & 0.984 & 0.997 & 0.990 & 0.947 & 0.973 & 566 & 9 & 2 \\
CSP   & 0.700 & 0.895 & 0.786 & 0.722 & 0.836 & 154 & 66 & 18 \\
LV    & 0.789 & 0.741 & 0.764 & 0.693 & 0.816 & 157 & 42 & 55 \\
\bottomrule
\end{tabular}
\end{table}

\begin{table}[!htb]
\centering
\caption{Per-class results, FetSAM Loss (YOLO26N, BCE-Dice-Lov\'{a}sz only).}
\label{tab:perclass_fetsam}
\setlength{\tabcolsep}{3pt}
\begin{tabular}{lcccccrrr}
\toprule
Class & Prec. & Rec. & F1 & IoU & DSC & TP & FP & FN \\
\midrule
Brain & 0.983 & 0.998 & 0.990 & 0.946 & 0.972 & 567 & 10 & 1 \\
CSP   & 0.640 & 0.837 & 0.725 & 0.715 & 0.831 & 144 & 81 & 28 \\
LV    & 0.655 & 0.618 & 0.636 & 0.712 & 0.828 & 131 & 69 & 81 \\
\bottomrule
\end{tabular}
\end{table}

\begin{figure}[!htb]
\centering
\includegraphics[width=\columnwidth]{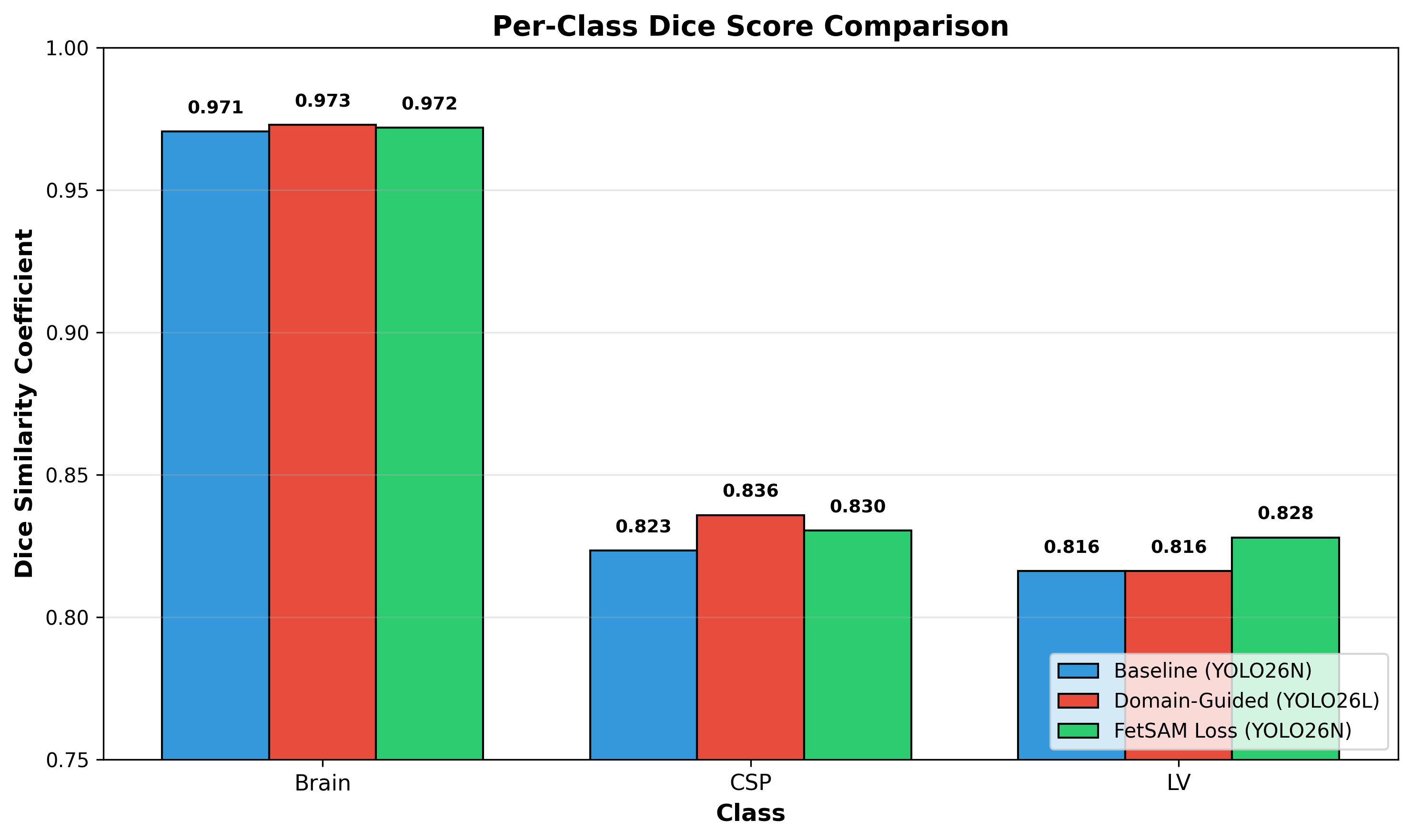}
\caption{Per-class DSC across the three models, with numeric labels.}
\label{fig:perclass_dice}
\end{figure}

\begin{figure}[!htb]
\centering
\includegraphics[width=\columnwidth]{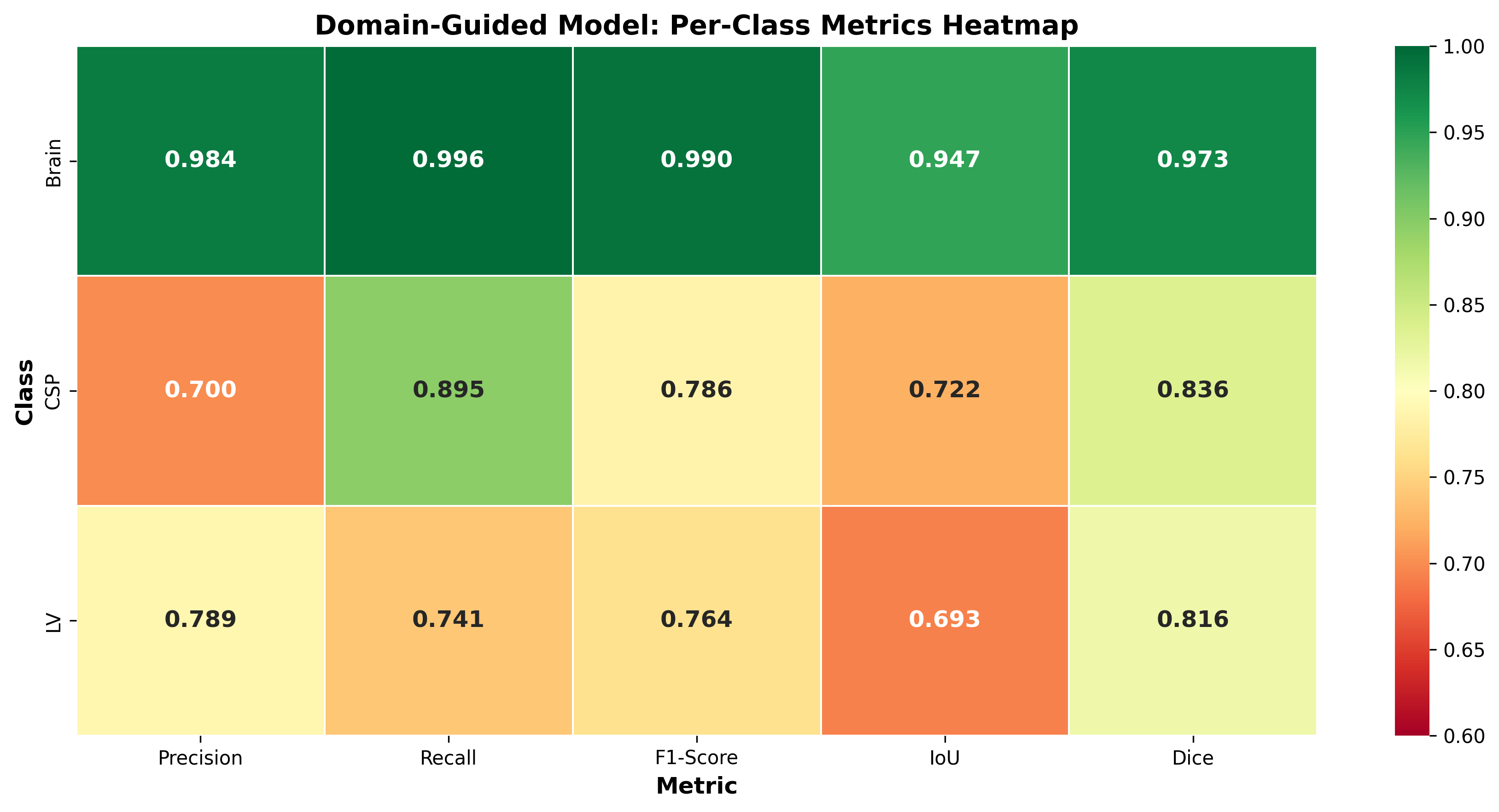}
\caption{Metric heatmap for the Domain-Guided model; warm colours highlight the Brain--CSP--LV performance disparity.}
\label{fig:heatmap}
\end{figure}

\Cref{tab:perclass_baseline,tab:perclass_dg,tab:perclass_fetsam} report the per-class detection and
segmentation metrics for each configuration. Brain DSC sits above 0.97 in all three runs. CSP and LV
hover in the 0.82--0.84 range, roughly 15 points lower. The gap is discussed in \Cref{sec:discussion}.
\Cref{fig:perclass_dice} compares per-class DSC across the three models, and \Cref{fig:heatmap} presents
a metric heatmap for the Domain-Guided model, highlighting the Brain--CSP--LV performance disparity.

\subsection{Training Dynamics}

\begin{figure}[!htb]
\centering
\includegraphics[width=\columnwidth]{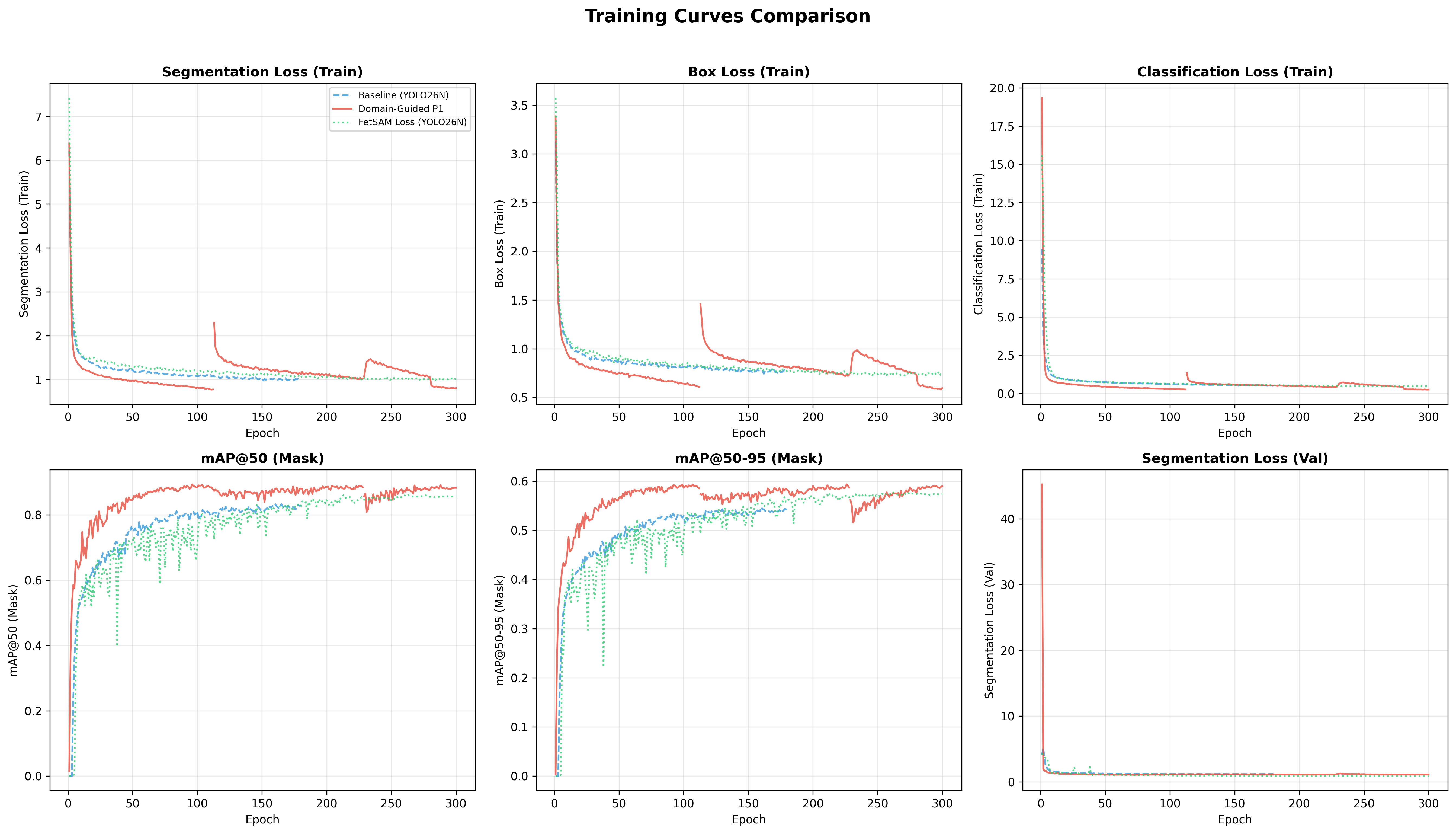}
\caption{Training curves: segmentation loss, box loss, classification loss, mAP@50, mAP@50-95, and validation segmentation loss versus epoch.}
\label{fig:training_curves}
\end{figure}

\begin{figure}[!htb]
\centering
\includegraphics[width=\columnwidth]{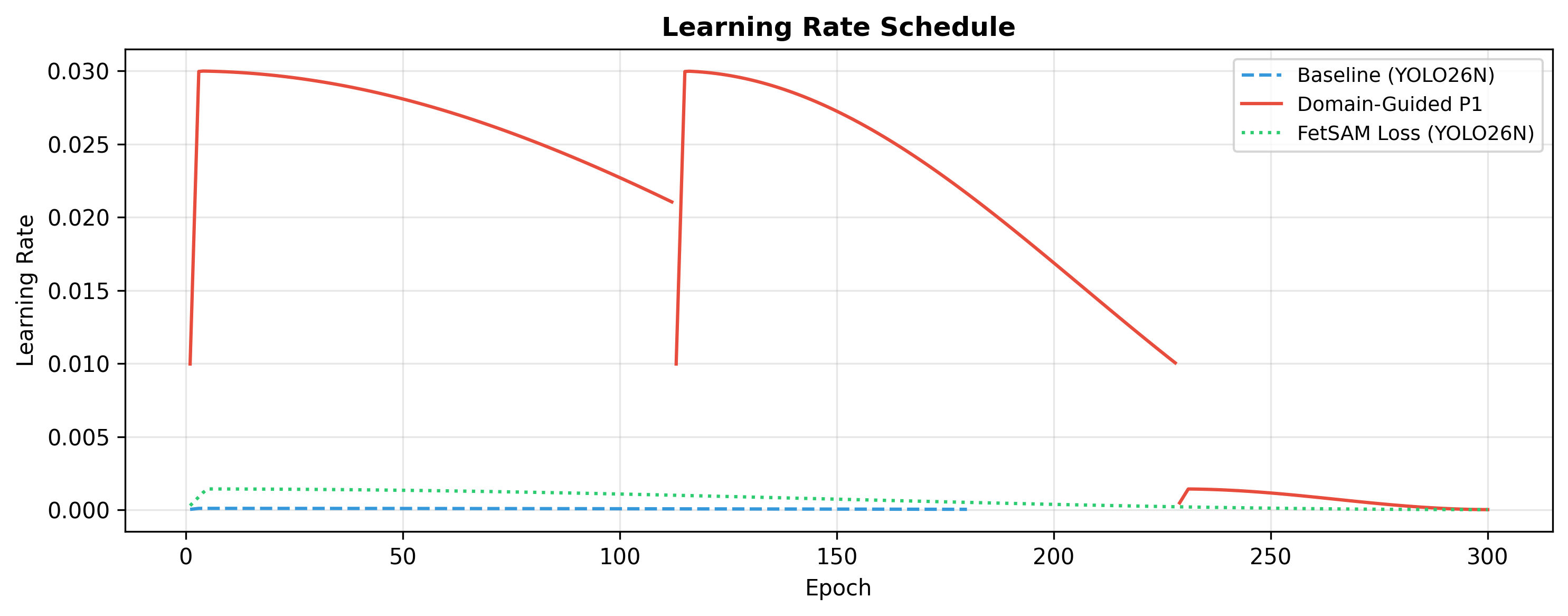}
\caption{Learning-rate schedules. All runs use cosine annealing with warm restarts~\cite{loshchilov2017sgdr}; the Domain-Guided curve resets at session boundaries.}
\label{fig:lr_schedule}
\end{figure}

Three observations stand out (\Cref{fig:training_curves}). First, the Baseline converges fast and triggers
early stopping at epoch~180 (patience~20); mAP is still creeping upward at that point. Second, the
Domain-Guided model, trained across three sessions with MuSGD, keeps improving through epoch~300,
suggesting it had not fully plateaued. Third, the FetSAM Loss run (same YOLO26N as Baseline, different
loss) reaches lower validation segmentation loss throughout, confirming that the Lov\'{a}sz term provides
gradient signal that BCE+Dice alone does not. \Cref{fig:lr_schedule} shows the learning-rate schedules
for all three runs.

\subsection{Comparison with FetSAM and Other Models}

\Cref{tab:comparison} compares mean DSC across all models, and \Cref{tab:perclass_fetsam_comparison}
provides a per-class DSC breakdown against FetSAM.

\begin{table}[!htb]
\centering
\caption{Mean DSC comparison with models reported in Alzubaidi et~al.~\cite{alzubaidi2024}.}
\label{tab:comparison}
\setlength{\tabcolsep}{2.5pt}
\begin{tabular}{llcr}
\toprule
Model & Backbone & Paradigm & Mean DSC \\
\midrule
FetSAM~\cite{alzubaidi2024} & SAM (ViT) & Prompt-based & 0.9012 \\
Efb0\_DLabV3+~\cite{alzubaidi2024} & EffNet-B0~\cite{tan2019} & Semantic & 0.8532 \\
Mit\_SegFormer\_b0~\cite{alzubaidi2024} & MiT-B0~\cite{xie2021} & Semantic & 0.8323 \\
Efb0\_Unet~\cite{alzubaidi2024} & EffNet-B0~\cite{tan2019} & Semantic & 0.8466 \\
Efb0\_Unet++~\cite{alzubaidi2024} & EffNet-B0~\cite{tan2019} & Semantic & 0.8495 \\
Efb0\_Linknet~\cite{alzubaidi2024} & EffNet-B0~\cite{tan2019} & Semantic & 0.8184 \\
Efb0\_FPN~\cite{alzubaidi2024} & EffNet-B0~\cite{tan2019} & Semantic & 0.8375 \\
Efb0\_PAN~\cite{alzubaidi2024} & EffNet-B0~\cite{tan2019} & Semantic & 0.8450 \\
Mvit075\_Unet~\cite{alzubaidi2024} & MobileViT-0.75 & Semantic & 0.7962 \\
Mvit075\_DLabV3~\cite{alzubaidi2024} & MobileViT-0.75 & Semantic & 0.8028 \\
Mvit075\_FPN~\cite{alzubaidi2024} & MobileViT-0.75 & Semantic & 0.7847 \\
Mvit075\_Manet~\cite{alzubaidi2024} & MobileViT-0.75 & Semantic & 0.7996 \\
Mvit075\_PSPNet~\cite{alzubaidi2024} & MobileViT-0.75 & Semantic & 0.7914 \\
\midrule
\textbf{Ours, FetSAM Loss} & YOLO26N & Instance & \textbf{0.9253} \\
\textbf{Ours, Baseline} & YOLO26N & Instance & \textbf{0.9229} \\
\textbf{Ours, Domain-Guided} & YOLO26L & Instance & \textbf{0.9208} \\
\bottomrule
\end{tabular}
\end{table}

All three of our runs surpass FetSAM's reported 0.9012. FetSAM's ``mean'' includes a background class
(BG DSC $\approx$ 0.9951), which inflates the average; ours is computed over the three foreground classes
only. Among semantic segmentation baselines, the EfficientNet-B0~\cite{tan2019} and
SegFormer~\cite{xie2021} backbones paired with various decoder architectures, U-Net~\cite{ronneberger2015},
UNet++~\cite{zhou2018}, DeepLabV3+~\cite{chen2018deeplabv3p}, FPN~\cite{lin2017fpn}, and
PAN~\cite{liu2018panet}, all achieve mean DSCs between 0.78 and 0.85, confirming that the YOLO26-based
instance segmentation approach provides a clear advantage.

\begin{table}[!htb]
\centering
\caption{Per-class DSC comparison with FetSAM~\cite{alzubaidi2024}.}
\label{tab:perclass_fetsam_comparison}
\begin{tabular}{lccc}
\toprule
Class & FetSAM & Ours (FetSAM Loss) & Ours (Dom.-Guided) \\
\midrule
Brain & 0.9851 & 0.9718 & 0.9729 \\
CSP   & 0.8037 & 0.8305 ($+$3.3\%) & \textbf{0.8359} ($+$4.0\%) \\
LV    & 0.8208 & \textbf{0.8280} ($+$0.9\%) & 0.8162 \\
Mean (fg) & 0.9012* & \textbf{0.9253} & 0.9208 \\
\bottomrule
\multicolumn{4}{l}{\footnotesize *FetSAM's published mean includes the background class.}
\end{tabular}
\end{table}

\section{Discussion}
\label{sec:discussion}

\subsection{Annotation Quality}

\begin{figure}[!htb]
\centering
\includegraphics[width=\columnwidth]{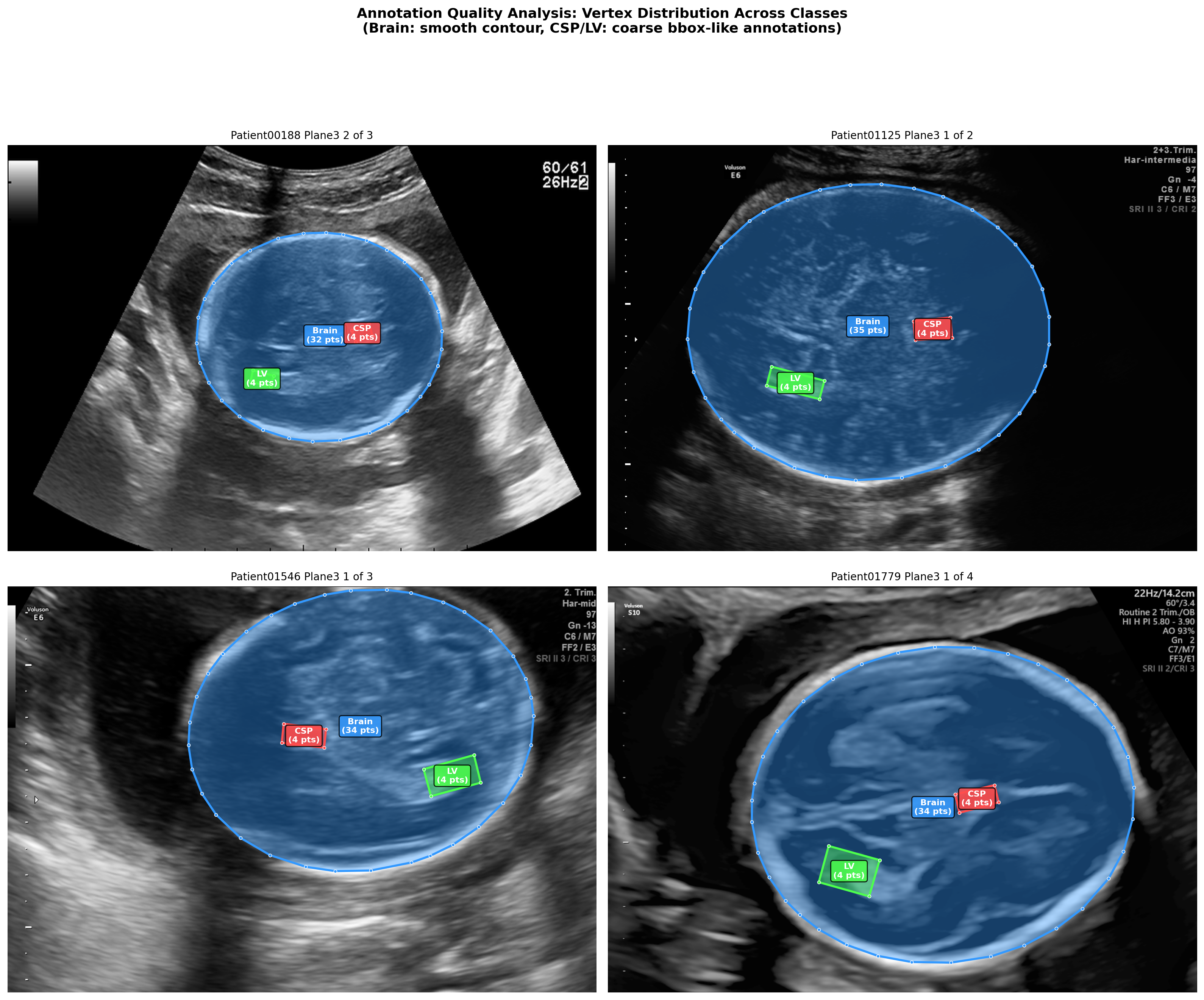}
\caption{Vertex counts per annotation. Brain contours (often $>$50 vertices) are substantially more detailed than CSP and LV masks (frequently 4--8 vertices).}
\label{fig:annotation_quality}
\end{figure}

\begin{figure}[!htb]
\centering
\includegraphics[width=\columnwidth]{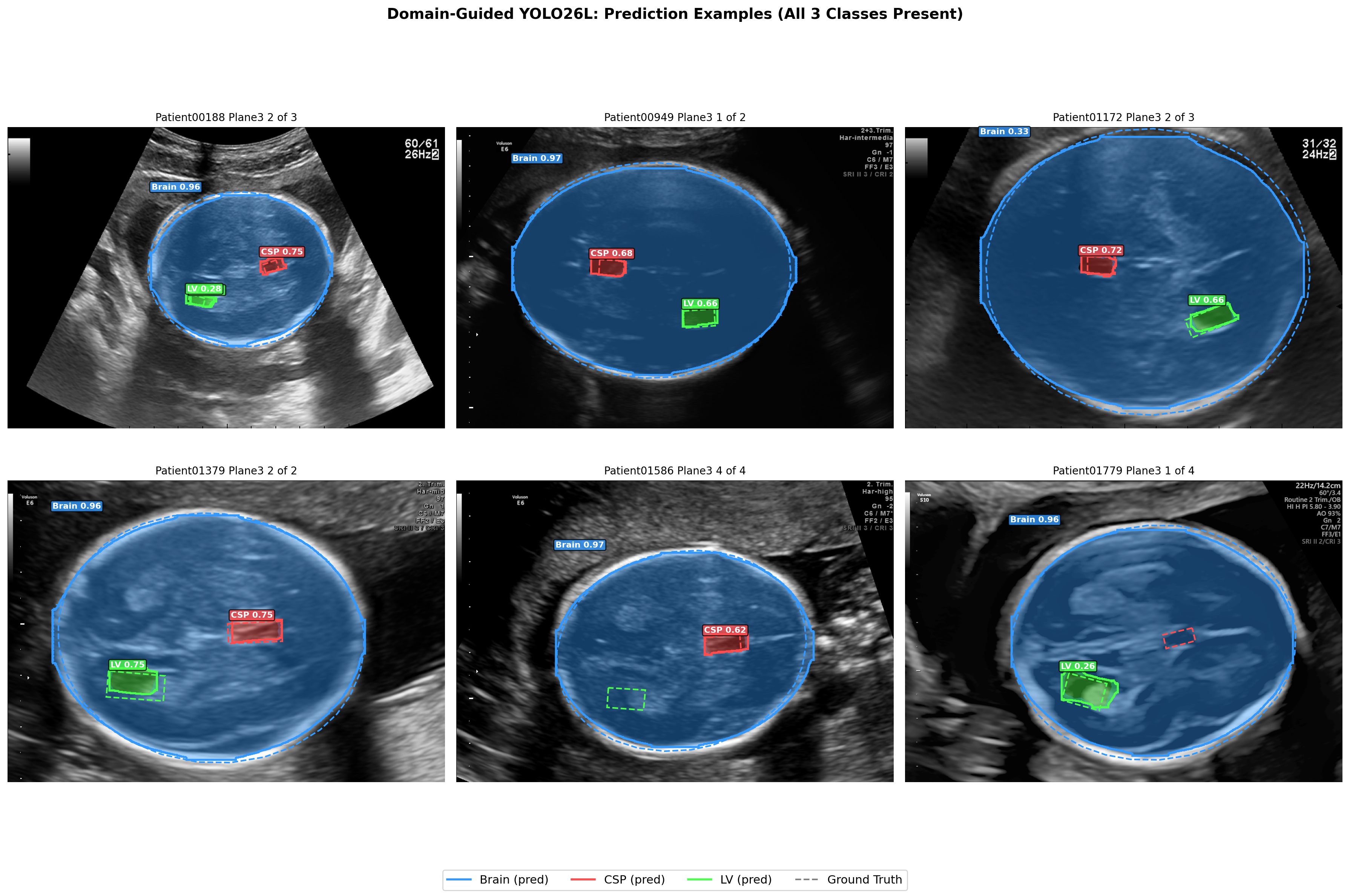}
\caption{Model predictions (contours) versus ground truth (filled) on test images containing all three classes. The model often produces smoother boundaries than the coarsely annotated references.}
\label{fig:predictions}
\end{figure}

Ground-truth masks for CSP and LV often consist of very few polygon vertices, sometimes as few as
four, yielding rectangular approximations rather than anatomically faithful contours. This coarseness
degrades the training signal (the network learns to match rectangles, not anatomy) and complicates metric
interpretation (both prediction and reference may be similarly rough, so DSC appears higher than the
actual boundary quality warrants). \Cref{fig:annotation_quality,fig:predictions} illustrate this issue.

\subsection{Composite Loss Ablation}

Holding architecture and data fixed (both YOLO26N, same dataset), swapping BCEDice for
BCE-Dice-Lov\'{a}sz yields $+$3.7\,pp mAP@50, $+$6.5\,pp mAP@50-95, and $+$0.24\,pp mDSC. The benefit
concentrates on LV, whose DSC rises from 0.8163 to 0.8280 ($+$1.4\%) and IoU from 0.6947 to 0.7117
($+$2.4\%). This is consistent with the Lov\'{a}sz loss's property of directly optimising the Jaccard
index~\cite{berman2018}: for small structures, even a few misclassified pixels cause a large relative
IoU drop, and the Lov\'{a}sz gradient penalises exactly that scenario.

\subsection{Model Capacity and Domain-Guided Augmentation}

The Domain-Guided run bundles three changes, larger backbone, composite loss, and augmented data, so its
gains cannot be attributed to a single factor. Taken together, they push mAP@50 from 0.7423 to 0.8657
($+$16.6\,pp) and Macro F1 from 0.7573 to 0.8467 ($+$11.8\,pp). Most of this improvement is in
\emph{detection}: CSP false negatives drop from 41 to 18 ($-$56\%) and LV false negatives from 85 to 55
($-$35\%). Per-instance segmentation quality barely changes (mDSC 0.9229 $\to$ 0.9208), suggesting the
extra capacity and augmented training data help the model \emph{find} minority structures, while the mask
quality per detected instance is already near the ceiling imposed by annotation granularity.

\subsection{Class Imbalance}

A 15-point DSC gap between Brain (0.97) and CSP/LV (0.82--0.84) persists in every configuration. Two
factors interact: (i)~Brain is large enough that even an imprecise mask overlaps substantially with the
ground truth, whereas CSP and LV are small and demand pixel-accurate boundaries; (ii)~training sees
roughly 3$\times$ more Brain instances, so the network's implicit prior favours the majority class.

Possible remedies include instance-level oversampling of CSP/LV, curriculum learning that gradually shifts
focus toward the harder classes, cropped-region training at higher resolution around small structures,
and, most directly, re-annotating CSP/LV with denser polygon vertices.

\subsection{Practical Considerations}

YOLO26N-Seg achieves an inference throughput of approximately 476 frames/s on a T4 GPU with TensorRT~10
optimization; the larger YOLO26L variant reaches about 120 frames/s. Neither requires bounding-box
prompts, which removes a step that FetSAM-style pipelines need at inference time. The prompt-free design
simplifies deployment in clinical settings where radiologist interaction should be minimal.

\section{Conclusion}
\label{sec:conclusion}

We have presented a fully automatic fetal head segmentation pipeline built on YOLO26-Seg with three
practical enhancements: a composite BCE-Dice-Lov\'{a}sz loss, domain-guided copy-paste augmentation, and
inter-patient data splitting. The composite loss alone raises mean DSC to 0.9253, outperforming FetSAM's
0.9012 by 2.68\,pp on a stricter three-class (no background) basis. Domain-guided augmentation provides
the largest detection improvement, cutting CSP false negatives by 56\% and LV false negatives by 35\%.
The remaining bottleneck is annotation quality: coarsely drawn CSP and LV masks limit both training signal
and achievable metric scores. Re-annotation with finer polygons is the most direct path to further gains.

All data and code are publicly available.


\appendices
\section{Reproducibility}

\begin{itemize}
  \item \textbf{Code:} \githuburl
  \item \textbf{Inter-Patient Split Dataset:} \kaggleurl
\end{itemize}

\end{document}